\documentstyle[makeidx,editedvolume]{crckapb}
\makeindex
\makeatletter
\def\cite{\def\citename##1{##1, }(\@internalcite}
\def\namecite{\def\citename##1{##1 (}\@internalcite}
\def\@cite#1#2{{#1\if@tempswa , #2\fi})}
\makeatother

\catcode`\@=11\relax
\newwrite\@unused
\def\typeout#1{{\let\protect\string\immediate\write\@unused{#1}}}
\def\figurepath{./}

\def\@nnil{\@nil}
\def\@empty{}
\def\@psdonoop#1\@@#2#3{}
\def\@psdo#1:=#2\do#3{\edef\@psdotmp{#2}\ifx\@psdotmp\@empty \else
    \expandafter\@psdoloop#2,\@nil,\@nil\@@#1{#3}\fi}
\def\@psdoloop#1,#2,#3\@@#4#5{\def#4{#1}\ifx #4\@nnil \else
       #5\def#4{#2}\ifx #4\@nnil \else#5\@ipsdoloop #3\@@#4{#5}\fi\fi}
\def\@ipsdoloop#1,#2\@@#3#4{\def#3{#1}\ifx #3\@nnil 
       \let\@nextwhile=\@psdonoop \else
      #4\relax\let\@nextwhile=\@ipsdoloop\fi\@nextwhile#2\@@#3{#4}}
\def\@tpsdo#1:=#2\do#3{\xdef\@psdotmp{#2}\ifx\@psdotmp\@empty \else
    \@tpsdoloop#2\@nil\@nil\@@#1{#3}\fi}
\def\@tpsdoloop#1#2\@@#3#4{\def#3{#1}\ifx #3\@nnil 
       \let\@nextwhile=\@psdonoop \else
      #4\relax\let\@nextwhile=\@tpsdoloop\fi\@nextwhile#2\@@#3{#4}}
\def\psdraft{
	\def\@psdraft{0}
}
\def\psfull{
	\def\@psdraft{100}
}
\psfull
\newif\if@prologfile
\newif\if@postlogfile
\newif\if@noisy
\def\pssilent{
	\@noisyfalse
}
\def\psnoisy{
	\@noisytrue
}
\psnoisy
\newif\if@bbllx
\newif\if@bblly
\newif\if@bburx
\newif\if@bbury
\newif\if@height
\newif\if@width
\newif\if@scale
\newif\if@rheight
\newif\if@rwidth
\newif\if@clip
\newif\if@verbose
\def\@p@@sclip#1{\@cliptrue}

\def\@p@@sfile#1{\def\@p@sfile{null}%
	        \openin1=#1
		\ifeof1\closein1%
		       \openin1=\figurepath#1
			\ifeof1\typeout{Error, File #1 not found}
			\else\closein1
			    \edef\@p@sfile{\figurepath#1}%
                        \fi%
		 \else\closein1%
		       \def\@p@sfile{#1}%
		 \fi}
\def\@p@@sfigure#1{\def\@p@sfile{null}%
	        \openin1=#1
		\ifeof1\closein1%
		       \openin1=\figurepath#1
			\ifeof1\typeout{Error, File #1 not found}
			\else\closein1
			    \def\@p@sfile{\figurepath#1}%
                        \fi%
		 \else\closein1%
		       \def\@p@sfile{#1}%
		 \fi}

\def\@p@@sbbllx#1{
		\@bbllxtrue
		\dimen100=#1
		\edef\@p@sbbllx{\number\dimen100}
}
\def\@p@@sbblly#1{
		\@bbllytrue
		\dimen100=#1
		\edef\@p@sbblly{\number\dimen100}
}
\def\@p@@sbburx#1{
		\@bburxtrue
		\dimen100=#1
		\edef\@p@sbburx{\number\dimen100}
}
\def\@p@@sbbury#1{
		\@bburytrue
		\dimen100=#1
		\edef\@p@sbbury{\number\dimen100}
}
\def\@p@@sscale#1{
		\@scaletrue
		\count255=#1
   		\edef\@p@sscale{\number\count255}
}
\def\@p@@sheight#1{
		\@heighttrue
		\dimen100=#1
   		\edef\@p@sheight{\number\dimen100}
}
\def\@p@@swidth#1{
		\@widthtrue
		\dimen100=#1
		\edef\@p@swidth{\number\dimen100}
}
\def\@p@@srheight#1{
		\@rheighttrue
		\dimen100=#1
		\edef\@p@srheight{\number\dimen100}
}
\def\@p@@srwidth#1{
		\@rwidthtrue
		\dimen100=#1
		\edef\@p@srwidth{\number\dimen100}
}
\def\@p@@ssilent#1{ 
		\@verbosefalse
}
\def\@p@@sprolog#1{\@prologfiletrue\def\@prologfileval{#1}}
\def\@p@@spostlog#1{\@postlogfiletrue\def\@postlogfileval{#1}}
\def\@cs@name#1{\csname #1\endcsname}
\def\@setparms#1=#2,{\@cs@name{@p@@s#1}{#2}}
\def\ps@init@parms{
		\@bbllxfalse \@bbllyfalse
		\@bburxfalse \@bburyfalse
		\@heightfalse \@widthfalse
		\@scalefalse
		\@rheightfalse \@rwidthfalse
		\def\@p@sbbllx{}\def\@p@sbblly{}
		\def\@p@sbburx{}\def\@p@sbbury{}
		\def\@p@sheight{}\def\@p@swidth{}
		\def\@p@sscale{}
		\def\@p@srheight{}\def\@p@srwidth{}
		\def\@p@sfile{}
		\def\@p@scost{10}
		\def\@sc{}
		\@prologfilefalse
		\@postlogfilefalse
		\@clipfalse
		\if@noisy
			\@verbosetrue
		\else
			\@verbosefalse
		\fi
}
\def\parse@ps@parms#1{
	 	\@psdo\@psfiga:=#1\do
		   {\expandafter\@setparms\@psfiga,}}
\newif\ifno@bb
\newif\ifnot@eof
\newread\ps@stream
\def\bb@missing{
	\if@verbose{
		\typeout{psfig: searching \@p@sfile \space  for bounding box}
	}\fi
	\openin\ps@stream=\@p@sfile
	\no@bbtrue
	\not@eoftrue
	\catcode`\%=12
	\loop
		\read\ps@stream to \line@in
		\global\toks200=\expandafter{\line@in}
		\ifeof\ps@stream \not@eoffalse \fi
		\@bbtest{\toks200}
		\if@bbmatch\not@eoffalse\expandafter\bb@cull\the\toks200\fi
	\ifnot@eof \repeat
	\catcode`\%=14
}	
\catcode`\%=12
\newif\if@bbmatch
\def\@bbtest#1{\expandafter\@a@\the#1
\long\def\@a@#1
\long\def\bb@cull#1 #2 #3 #4 #5 {
	\dimen100=#2 bp\edef\@p@sbbllx{\number\dimen100}
	\dimen100=#3 bp\edef\@p@sbblly{\number\dimen100}
	\dimen100=#4 bp\edef\@p@sbburx{\number\dimen100}
	\dimen100=#5 bp\edef\@p@sbbury{\number\dimen100}
	\no@bbfalse
}
\catcode`\%=14
\def\compute@bb{
		\no@bbfalse
		\if@bbllx \else \no@bbtrue \fi
		\if@bblly \else \no@bbtrue \fi
		\if@bburx \else \no@bbtrue \fi
		\if@bbury \else \no@bbtrue \fi
		\ifno@bb \bb@missing \fi
		\ifno@bb \typeout{FATAL ERROR: no bb supplied or found}
			\no-bb-error
		\fi
		\count203=\@p@sbburx
		\count204=\@p@sbbury
		\advance\count203 by -\@p@sbbllx
		\advance\count204 by -\@p@sbblly
		\edef\@bbw{\number\count203}
		\edef\@bbh{\number\count204}
}
\def\in@hundreds#1#2#3{\count240=#2 \count241=#3
		     \count100=\count240	
		     \divide\count100 by \count241
		     \count101=\count100
		     \multiply\count101 by \count241
		     \advance\count240 by -\count101
		     \multiply\count240 by 10
		     \count101=\count240	
		     \divide\count101 by \count241
		     \count102=\count101
		     \multiply\count102 by \count241
		     \advance\count240 by -\count102
		     \multiply\count240 by 10
		     \count102=\count240	
		     \divide\count102 by \count241
		     \count200=#1\count205=0
		     \count201=\count200
			\multiply\count201 by \count100
		 	\advance\count205 by \count201
		     \count201=\count200
			\divide\count201 by 10
			\multiply\count201 by \count101
			\advance\count205 by \count201
		     \count201=\count200
			\divide\count201 by 100
			\multiply\count201 by \count102
			\advance\count205 by \count201
		     \edef\@result{\number\count205}
}
\def\compute@wfromh{
		\in@hundreds{\@p@sheight}{\@bbw}{\@bbh}
		\edef\@p@swidth{\@result}
}
\def\compute@hfromw{
		\in@hundreds{\@p@swidth}{\@bbh}{\@bbw}
		\edef\@p@sheight{\@result}
}
\def\compute@wfroms{
		\in@hundreds{\@p@sscale}{\@bbw}{100}
		\edef\@p@swidth{\@result}
}
\def\compute@hfroms{
		\in@hundreds{\@p@sscale}{\@bbh}{100}
		\edef\@p@sheight{\@result}
}
\def\compute@handw{
		\if@scale
			\compute@wfroms
			\compute@hfroms
		\else
			\if@height 
				\if@width
				\else
					\compute@wfromh
				\fi	
			\else 
				\if@width
					\compute@hfromw
				\else
					\edef\@p@sheight{\@bbh}
					\edef\@p@swidth{\@bbw}
				\fi
			\fi
		\fi
}
\def\compute@resv{
		\if@rheight \else \edef\@p@srheight{\@p@sheight} \fi
		\if@rwidth \else \edef\@p@srwidth{\@p@swidth} \fi
}
\def\compute@sizes{
	\compute@bb
	\compute@handw
	\compute@resv
}
\def\psfig#1{\vbox {
	\ps@init@parms
	\parse@ps@parms{#1}
	\compute@sizes
	\ifnum\@p@scost<\@psdraft{
		\if@verbose{
			\typeout{psfig: including \@p@sfile \space }
		}\fi
		\special{ps::[begin] 	\@p@swidth \space \@p@sheight \space
				\@p@sbbllx \space \@p@sbblly \space
				\@p@sbburx \space \@p@sbbury \space
				startTexFig \space }
		\if@clip{
			\if@verbose{
				\typeout{(clip)}
			}\fi
			\special{ps:: doclip \space }
		}\fi
		\if@prologfile
		    \special{ps: plotfile \@prologfileval \space } \fi
		\special{ps: plotfile \@p@sfile \space }
		\if@postlogfile
		    \special{ps: plotfile \@postlogfileval \space } \fi
		\special{ps::[end] endTexFig \space }
		\vbox to \@p@srheight true sp{
			\hbox to \@p@srwidth true sp{
				\hss
			}
		\vss
		}
	}\else{
		\vbox to \@p@srheight true sp{
		\vss
			\hbox to \@p@srwidth true sp{
				\hss
				\if@verbose{
					\@p@sfile
				}\fi
				\hss
			}
		\vss
		}
	}\fi
}}
\def\psglobal{\typeout{psfig: PSGLOBAL is OBSOLETE; use psprint -m instead}}
\catcode`\@=12\relax

\begin{opening}
\title{TAGGING FRENCH WITHOUT LEXICAL\protect\\
PROBABILITIES -- COMBINING LINGUISTIC\protect\\
KNOWLEDGE AND STATISTICAL LEARNING}

\author{Evelyne Tzoukermann}
\institute{AT\&T Bell Laboratories\\
600 Mountain Avenue\\
Murray Hill, NJ 07974--0636 \\
{\tt evelyne@research.att.com}}
\author{Dragomir R. Radev$^*$}
\institute{Department of Computer Science \\
450 Computer Science Building, Columbia University \\
New York, NY 10027\\
{\tt radev@cs.columbia.edu}}
\author{William A. Gale}
\institute{AT\&T Bell Laboratories\\
600 Mountain Avenue\\
Murray Hill, NJ 07974--0636 \\
{\tt gale@research.att.com}}
\end{opening}

\runningtitle{TAGGING FRENCH WITHOUT LEXICAL PROBABILITIES}

\begin{document}
\renewcommand{\thefootnote}{\fnsymbol{footnote}}
\footnotetext[1]{The work was achieved while the author was at
AT\&T Bell Laboratories, 600 Mountain Avenue,
Murray Hill, NJ 07974--0636}
\renewcommand{\thefootnote}{\arabic{footnote}}
\begin{abstract}
This paper explores morpho-syntactic ambiguities\index{morpho-syntactic ambiguities}
for French\index{French} to develop
a strategy for part-of-speech disambiguation\index{part-of-speech disambiguation}
that a) reflects the
complexity of French as an inflected language, b) optimizes
the estimation of probabilities, c) allows the user flexibility in
choosing a tagset.
The problem in extracting lexical probabilities\index{lexical probabilities}
from a limited training corpus is that the statistical model may not
necessarily represent the use of a particular word in a particular
context.  In a highly morphologically inflected language, this
argument is particularly serious since a word can be tagged with a
large number of parts of speech\index{parts-of-speech}.
Due to the lack of sufficient training data, we
argue against estimating
lexical probabilities\index{lexical probabilities} to disambiguate parts
of speech in unrestricted texts.  Instead, we use the strength of
contextual probabilities\index{contextual probabilities} along with a feature we call 
``genotype\index{genotype}'', a set of tags associated with a word.  Using this
knowledge, we have built a part-of-speech tagger\index{part-of-speech tagger}
that combines
linguistic and statistical approaches: contextual information is
disambiguated by linguistic rules and {\it n}-gram probabilities on
parts of speech only are estimated in order to disambiguate the remaining
ambiguous tags.
\end{abstract}

\section{Introduction}
This paper explores morpho-syntactic ambiguities\index{morpho-syntactic ambiguities}
for French\index{French} to develop
a strategy to disambiguate part of speech\index{part-of-speech}
labels that a) reflects the
nature of French as an inflected language, b) optimizes the estimation
of probabilities, c) allows the user flexibility in tagging.\bigskip\\
{\bf Problems in tagging French:}\\ French\index{French} has a rich
inflectional morphology\index{inflectional morphology}
and words can have up to eight different
morphological analysis depending on the choice of tags.  For example, let us take a
common word of French, the word ``moyenne'' (meaning {\it average} as
a noun, verb, or adjective) shown in Table~\ref{marine}. The word has
seven distinct morphological analyses\index{morphological analyses}.
Column 3 gives the full
morphological analysis of the word, column 4 represents the tag
corresponding to it from the large tagset\index{tagset} and column 5, the tag from
the small tagset\index{tagset} (large and small tagsets are discussed in
Sections~\ref{tagset} and \ref{comp}).  The seven tags in the large
tagset get reduced to five in the small one.

\begin{table}[htbp] 
\footnotesize
\centering
\begin{tabular}{|l|l|l|l|l|} \hline
{\bf word} & {\bf base form} & {\bf morphological analysis} & {\bf
tagset1} & {\bf tagset2}\\ \hline
  ``moyenne'' & $<$moyen$>$   &  adjective, fem. sing. & JFS & jfs\\
  ``moyenne'' & $<$moyenne$>$  &  noun, feminine sing. & NFS & nfs\\
  ``moyenne'' & $<$moyenner$>$ &  verb, 1st pers., sing., pres.,
ind. & V1SPI & v1s\\
  ``moyenne'' & $<$moyenner$>$ &  verb, 1st pers., sing., pres.,
subj. & V1SPS & v1s\\
  ``moyenne'' & $<$moyenner$>$ &  verb, 2nd pers., sing., pres.,
imp. & V2SPM & v2s\\
  ``moyenne'' & $<$moyenner$>$ &  verb, 3rd pers., sing., pres.,
ind. & V3SPI & v3s\\
  ``moyenne'' & $<$moyenner$>$ &  verb, 3rd pers., sing., pres.,
subj. & V3SPS & v3s\\ \hline
\end{tabular} 
\caption{Morphological analyses of the word ``moyenne''. \label{marine}}
\end{table}

In a given sentence where the word ``moyenne'' occurs, 
multiple other tags appear, as exemplified in Table~\ref{morph}.
The second column of the table shows all the tags for the word in column 1.
The correct tag is in bold, followed by the meaning of the correct tag in column 3.

\begin{table}[htbp]
\small
\centering
\begin{tabular}{|l|l|l|} \hline
{\bf Word} & {\bf tag from morphology} & {\bf Meaning of the tag} \\ \hline
$<S>$          &   \^{ } &  beginning of sentence  \\         
La             &  {\bf rf} b nms u   &  article  \\         
teneur         &   {\bf nfs} nms   &  noun feminine singular  \\         
{\bf moyenne}  &   {\bf jfs} nfs v1s v2s v3s  &  {\bf adjective feminine singular}  \\         
en             &   {\bf p} a b    &  preposition  \\         
uranium        &   {\bf nms}   &  noun masculine singular  \\         
des            &   {\bf p} r    &   preposition \\         
rivi\`eres      &   {\bf nfp}   &  noun feminine plural  \\         
,              &   {\bf x}     &   punctuation \\         
bien\_que       &   {\bf cs}    &  subordinating conjunction  \\         
d\'elicate      &   {\bf jfs}   &  adjective feminine singular   \\         
\`a             &   {\bf p}     &  preposition  \\         
calculer       &   {\bf v}     &   verb \\    \hline     
\end{tabular} 
\normalsize
\caption{Sample output of a sentence chunk with the
word ``moyenne''. \label{morph}}
\end{table}

\normalsize 
The goal of tagging is to find the most appropriate tag
associated with a word. It has often been suggested that
lexical probabilities\index{lexical probabilities}
should be used on word forms in order to find the most
likely tag for a word. This approach is somewhat limited for tagging
richly inflected languages, especially when in addition to the part of
speech, the output of the system needs to contain morphological
information (such as number, tense, and person). The problem with
extracting lexical probabilities\index{lexical probabilities}
from a limited training corpus is
related to the fact that statistics may not necessarily represent the use of a particular word in a particular context. In French\index{French}, a word
can have up to eight parts of speech, and it is very unlikely that all
corresponding forms will be present in the training corpus in large
enough numbers.

Our goal is to identify approaches that allow for a better estimation of the
variability of tag distributions\index{tag distributions}
for all words that appear in the test
corpus.  Several paradigms have been used for disambiguating parts of speech in
French. Whether one or another should be used depends on the availability
of large training corpora as well as on the amount of information that the
tags are used to convey.
The next section explores different strategies to handle the
morphological variability of French\index{morphological variability of French},
and proposes a solution which captures
variability on one hand, and frequency of patterns on the other.
Section~\ref{lex-cont} gives some evidence on the power of
contextual probabilities\index{contextual probabilities} vs.
lexical\index{lexical probabilities} ones for French.  Finally, the paper presents a
part of speech tagger that takes into account both
linguistic knowledge\index{linguistic knowledge} and
statistical\index{statistical knowledge} learning.
Its novelty relies on several features: (a) the estimation
of probabilities based on genotypes\index{genotype}, (b) a fully
modular architecture
that allows the user flexible ordering of all independent
modules, (c) an expanded tagset\index{tagset} that gives the
user the flexibility\index{tagset!tagset flexibility} to use
any derived subset, (d) the exportability of the system to other languages,
and (e) the use of a mixed linguistic and statistical approach.  Results
are provided, as well as directions for future use of the model.

\section{Strategies to capture morphological variants}
  Given that a word can have from two to eight
different morphological types (based only on six morphological categories,
such as syntactic category (noun, adjectives, verbs, etc.) and mood, tense, person, number, gender), an
important step in designing a tagger is to decide which features the tagset
should capture.  Then, given the multitude of morphological variants\index{morphological variants}
(one single French\index{French} verb can have up to 45 inflected forms),
what is the best way to optimize the training corpus?
It is clear that learning the distribution
of a large variety of tags is very difficult with sparse training input.
Morphological variants could be obtained via:
\begin{itemize}
\item {\bf base forms:} in Table~\ref{marine}, the word ``moyenne''
has three different base forms, the masculine adjective ``moyen'',
the feminine noun ``moyenne'', and
the verb ``moyenner''.  One way to capture these
morphological variants could be to take the paradigm of base forms and
to estimate probabilities on the different inflections.  For example, in
the word {\it moyenne}, one could estimate the probabilities of the
verbal base form ``moyenn-er'' by the frequency of occurrences of the following
endings {\sc 1st\--per\-son-sin\-gu\-lar\--pre\-sent-ind\-ic\-ati\-ve}, 
{\sc 1st\--per\-son\--sin\-gu\-lar\--pre\-sent\--
sub\-junc\-ti\-ve}, 
{\sc 2nd\--per\-son\--sin\-gu\-lar\--pre\-sent\--im\-pe\-ra\-ti\-ve}, 
{\sc 3rd-
per\-son\--sin\-gu\-lar\--pre\-sent\--ind\-ica\-ti\-ve}, 
{\sc 3rd\--per\-son\--sin\-gu\-lar-
pre\-sent-\-\-sub\-junc\-ti\-ve}.  This would
almost rule out forms such as {\sc
2nd-person-singular-present-imperative}, since imperative forms would
be less likely to occur in narrative texts than indicative forms\footnote{Of
course, this would also depend on the genre of the text; imperative
forms would be more frequent in cookbooks for example.}.  Also, 1st
person forms would be given lower probabilities, since they are less
likely to appear in news articles.

\item {\bf surface forms:} another way to capture the information
could be to estimate the lexical probabilities of the words in a text.
That is, for each word such as ``moyenne'', estimate the probability
of the word given the eight morphological distinct forms.  This would
necessitate an extremely large body of texts in order to cover all the
inflectional variations for a given word.  Taking into account that
there is no disambiguated corpus of that size for French, this
approach does not seem feasible.
\end{itemize}

Taking into account these previous points, we have used a new paradigm
to capture the inflection of a word on the one hand, and the analyses
associated to this word on the other.  We call a {\bf genotype}\index{genotype} the
set of tags that a given word inherits from the morphological
analysis.  
For example, the French\index{French} word ``le'' (meaning {\it the} or
the direct object {\it it, him}) has two parts of speech: BD3S
[{\sc personal-pronoun-direct-3rd-per\-son-sin\-gu\-lar}] and 
RDM [{\sc definite-masculine-article}].  Thus, its genotype\index{genotype} is the set
[BD3S RDM].  Similarly, the genotype\index{genotype} for the word ``moyenne'' is 
[JFS, NFS, V1SPI, V1SPS, V2SPM, V3SPI, V3SPS]
or [jfs, nfs, v1s, v2s, v3s], depending on the tagset (see Sections~\ref{tagset} and \ref{comp}
for a description of the tagsets).

Section~\ref{lex-prob} demonstrates that words falling in the same
genotype\index{genotype} have similar distributions of parts of
speech\index{distributions of parts of speech}.  We will also show
that using genotypes\index{genotype} for disambiguation reduces the
sparseness of training data.  In some sense, this is comparable to the
approach taken in Cutting et al. (1992).  In this approach, they use
the notion of word equivalence or ambiguity classes to describe words
belonging to the same part-of-speech categories.  In our work, the
whole algorithm bases estimations on genotype\index{genotype} only,
filtering down the ambiguities and resolving them with statistics.
Moreover, the estimation is achieved on a sequence of {\it n-}gram
genotypes\index{genotype}.  Also, the refinement that is contained in
our system reflects the real morphological
ambiguities\index{morphological ambiguities}, due to the rich nature
of the morphological output and the choice of tags.  There are three
main differences between their work and ours.  First, in their work,
the most common words are estimated individually and the less common
ones are put together in their respective ambiguity classes; in our
work, every word is equally treated by its respective
genotype\index{genotype}.  Second, in their work, ambiguity classes
can be marked with a preferred tag in order to help disambiguation
whereas in our work, there is no special annotation since words get
disambiguated through the sequential application of the modules.
Third, and perhaps the most important, in our system, the linguistic
and statistical estimations are entirely done on the
genotypes\index{genotype} only, regardless of the words.  Words are
not estimated given their individual of class categories;
genotypes\index{genotype} are estimated alone (unigram probabilities)
or in the context of other genotypes\index{genotype} (bi- and tri-gram
probabilities).

\section{Lexical Probabilities vs. Contextual Probabilities}  \label{lex-cont}
 There has been considerable discussion in the literature on part of
speech tagging as to whether lexical probabilities\index{lexical probabilities}
are more important for probability estimation than
contextual\index{contextual probabilities} ones, and whether they are
more difficult to obtain, given the nature of corpora and the
associated problem of sparse data.  On one hand, Church (1992)
claims that it is worth focusing on lexical probabilities\index{lexical probabilities}, since this
is the actual weakness of present taggers.  On the other hand,
Voutilainen \cite{KarVouHeiAnt95} argues that word ambiguities vary
widely in function of the specific text and genre.  He gives the
example of the word ``cover'' that can be either a noun or a verb.  He
shows that even in the large collection of genres gathered under the
Brown Corpus \cite{Francis82} and the LOB Corpus \cite{Johansson80},
the homograph ``cover'' is a noun in 40\% of the cases, and a verb in
the rest.  The same homograph extracted from a car maintenance manual,
{\it always} appears as a noun.  Several experiments were run to
figure out the types of ambiguities found in French and their
distribution. 
In the tagger for French, we argue that
contextual probabilities\index{contextual probabilities} are
in fact more important to estimate than lexical ones since a) there is
no large training corpus for French\index{French}, b) it would be nearly impossible
to get a corpus covering all French morphological inflected forms. As
Zipf's law predicts, even an arbitrary large training corpus would
still be missing many word forms, since that corpus would have a large
tail of words occurring very few times. Zipf's law holds even stronger
for French.

\subsection{How ambiguous is French?}
We selected two corpora\footnote{Extract of the French newspapers Le
Monde (Paris), September-October, 1989, January, 1990.  Articles
Nos. 1490 - 1875.}, one with 94,882 tokens and the other with 200,182
tokens, in order to account for the
morpho-syntactic ambiguity\index{morpho-syntactic ambiguity} of
French.  Table~\ref{tab:dist} shows the distribution of these
ambiguities for each French\index{French} token.  Columns 2 and 4 give the number of
words corresponding to the tags in column 1.  Column 3 and 5 show the
percentage of words per tags in the corpus.

\begin{table}[htbp]
\footnotesize
\centering
\begin{tabular}{|l||l|l||l|l|} \hline
{\bf genotype}\index{genotype}  & {\bf 94,882} & {\bf \% of the} & {\bf 200,182} & {\bf
\% of the} \\
{\bf size}  & {\bf tokens} & {\bf corpus} & {\bf tokens} & {\bf corpus} \\\hline
1 tag &   54570  &   57\% &  	   110843  &  58\%   \\             
2 tags &  24636   &  26\% & 	    50984   & 25\%  \\          
3 tags &  11058   &  11\% & 	    23239   & 11\%  \\          
4 tags &    634   &  .5\% & 	     3108   & 1\%  \\          
5 tags &    856   &  .9\% & 	     5963   & 2\%  \\          
6 tags &   2221   &  2\% & 	     4621   & 2\%  \\          
7 tags &    590   &  .5\% & 	     1069   & .5\%  \\          
8 tags &    317   &  .5\% & 	      355   & .1\% \\ \hline
\end{tabular} 
\caption{Ambiguity of French words in two corpora of different sizes. \label{tab:dist}}
\end{table}

It is interesting to point out that despite the fact that one corpus
is twice the size of the other, the distribution of the number of tags
per word is nearly the same.  Table~\ref{tab:dist} shows that a little more
than half of the words in French\index{French} texts are unambiguous, 25\% of the
words have two tags, 11\% of the words have three tags, and about 5\%
of the words have from four to eight tags.  Another way to quantify the
word ambiguity is that,
for the corpus of 94,882 tokens, there is a
total of 163,824 tags, which gives an average ambiguity factor of 1.72
per word.  Similarly, for the corpus of 200,182 tokens, there are
362,824 tags, which gives an ambiguity factor of 1.81 per word.

\subsection{Lexical probabilities vs. genotypes}\index{lexical probabilities}\index{genotype} \label{lex-prob}
In Table~\ref{lex1}, a few words belonging to a very frequent genotype\index{genotype}
[nfs v1s v2s v3s] (noun-\-fem\-ini\-ne-\-sin\-gu\-lar,
verb-\-1st-\-per\-son-\-sin\-gu\-lar,
verb-\-2nd-\-per\-son-\-sin\-gu\-lar,
verb-\-3rd-\-per\-son-\-sin\-gu\-lar) were extracted from the test
corpus and probabilities were estimated with the information from the
training corpora.  The table shows the words in the leftmost column;
the next three columns display the distribution in the three corpora
(C$_1$, C$_2$, C$_3$), with the number of occurrences found in the
training corpus (``occ'' in the table), the number of times the word
is tagged ``nfs'' and the number of times it is ``v3s''.  Note that
since these words were never ``v1s'' or ``v2s'' in the training
corpus, there is no account for these parts-of-speech.  Column 4 shows
the total for the three corpora.  Table~\ref{lex2} gives a total in percentage
of the occurences of  ``nfs'' and ``v3s'' in the training corpora. 
The sum of the 8 words is given followed, in the last line of the table, by 
the resolution of this genotype\index{genotype} throughout the entire training corpus.

\tiny
\begin{table}[htbp]
\centering
\begin{tabular}{|l|ccc|ccc|ccc|ccc|} \hline
& \multicolumn{3}{c|}{\bf Training C$_1$} & \multicolumn{3}{c|}{\bf
Training C$_2$} & \multicolumn{3}{c|}{\bf Training C$_3$} &
\multicolumn{3}{c|}{\bf Training C$_{1-3}$}\\& \multicolumn{3}{c|}{\bf 10K words} & \multicolumn{3}{c|}{\bf 30K
words} & \multicolumn{3}{c|}{\bf 36K words} & \multicolumn{3}{c|}{\bf 76K words} \\\hline
& {\bf occ} & {\bf nfs} & {\bf v3s}   & {\bf occ} & {\bf nfs} & {\bf v3s} &  {\bf occ} &  {\bf nfs} & {\bf v3s} &   {\bf occ} & {\bf nfs} & {\bf v3s} \\\hline
{\bf laisse}      &     1 &  0  & 1  &   0 &  0 &  0  &   0  & 0 &  0 &   1 &  0 &  1 \\
{\bf masse}       &     0 &  0  & 0  &  11 & 11 &  0  &   0  & 0 &  0 &  11 & 11 &  0 \\
{\bf t\^ache}     &     2 &  2  & 0  &   0 &  0 &  0  &   0  & 0 &  0 &   2 &  2 &  0 \\
{\bf lutte}       &     0 &  0  & 0  &   0 &  0 &  0  &   5  & 4 &  1 &   5 &  4 &  1 \\ 
{\bf forme}       &     3 &  3  & 0  &   61 &  57 & 4  &  1  & 1 &  0 &   65 &  61 &  4 \\
{\bf zone}        &     0 &  0  & 0  &   12 &  12 & 0  &  5  & 5 & 0 &    17 &  17 &  0 \\
{\bf danse}       &     0 &  0  & 0  &   0 &  0 &  0  &   0  & 0 &  0 &    0 &  0 &  0 \\
{\bf place}       &     4 &  4  & 0  &   10 &  10 & 0  &  12  & 12 & 0 &  26 &  26 &  0 \\ \hline 
{\bf Total}: & 10 &  9 &  1 &   94 & 90 &  4  &   23 &  22 &  1  & 127 & 121 &  6  \\ \hline
\end{tabular} 
\normalsize
\caption{Comparing frequencies of words vs. genotypes\index{genotype}. \label{lex1}}
\end{table}

\begin{table}[htbp]
\centering
\begin{tabular}{|l|r|r|} \hline
& {\bf  Total nfs} & {\bf Total v3s}\\\hline
{\bf laisse}         &        0.00 \% &  100.00 \% \\
{\bf masse}          &      100.00 \% &    0.00 \% \\
{\bf t\^ache}        &      100.00 \% &    0.00 \% \\
{\bf lutte}          &       80.00 \% &   20.00 \% \\
{\bf forme}          &       94.00 \% &   6.00 \% \\
{\bf zone}           &      100.00 \% &   0.00 \% \\
{\bf danse}          &       NO DATA &   NO DATA \\
{\bf place}          &       100.00 \% &   0.00 \% \\\hline
{\bf Total 8 words}: &        95.2 \% &    4.72 \% \\
{\bf Total genotype}\index{genotype}:&     89.15 \% &   10.85 \% \\\hline
\end{tabular} 
\normalsize
\caption{Total of ``nfs'' vs. ``v3s'' in Table 4.} \label{lex2}
\end{table}

\normalsize 
Tables~\ref{lex1} and \ref{lex2} show that, if we were to estimate
lexical probabilities\index{lexical probabilities},
there would not be any information for the word
``danse'' ({\it dance}),
since it does not appear in the training corpus.  On the other hand, in
capturing only the genotype\index{genotype} [nfs, v1s, v2s, v3s] for the word ``danse'', 
the information from the training corpus of 89.15\% ``nfs'', 10.85\%
``v3s'' will be applied and ``danse'' will be correctly assigned the 
``nfs'' tag.  In this case, genotypes\index{genotype} help with smoothing since the word
itself (known or not from the training data) is ignored, and its membership
in the genotype\index{genotype} supercategory is used instead.  Therefore, this strategy
drastically reduces the problem of sparse data\index{sparse data}.
In the Brown corpus, about
40,000 words appear five times or less \cite{church92}.  When low frequency
words occur with equal part-of-speech distribution\index{part-of-speech distribution},
it is hardly possible to pick the right part-of-speech.
Using genotypes\index{genotype} is a practical way to solve that problem.

\subsection{Distribution of genotypes}\index{genotype}
To exemplify further the advantage of using genotypes\index{genotype}, we measured their
distribution through the same three corpora.  Table~\ref{geno}
exhibits some convincing numbers: for a corpus of 10,006 tokens, there
are 219 different genotypes\index{genotype}, and for a corpus of 76,162 tokens, there
are 304 unique genotypes\index{genotype}.  In other words, while the corpus is
increased by 86\%, the number of different genotypes\index{genotype} increases by 27\% only.
Furthermore, the number of genotypes\index{genotype} to estimate remain very low,
since for a corpus of 76,162 tokens, genotypes\index{genotype} represent only 3\% of this number.

\begin{table}[htbp]
\centering
\begin{tabular}{|l|l|l|l|} \hline
{\bf corpora} & {\bf \# of tokens} & {\bf \# of words} & {\bf \# of 
genotypes}\index{genotype} \\\hline
{\bf Corpus$_1$} & 10006 & 2767 & 219   \\
{\bf Corpus$_2$} & 34636 & 4714 & 241 \\
{\bf Corpus$_3$} & 31520 & 5299 & 262 \\
{\bf Corpus$_{1-3}$} & 76162 &10090 & 304 \\\hline
\end{tabular} 
\caption{Genotype distribution.  \label{geno}}
\end{table}

\section{Construction of the Tagger\index{tagger}}
The tagger\index{tagger} is made of a series of stand-alone modular programs which can be
combined in many different ways to allow flexibility of the system.

\begin{figure}[htb]
\centerline{\psfig{figure=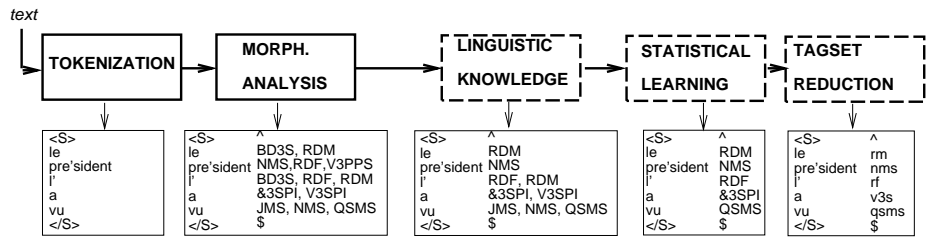,scale=57}}
\caption{System Components.  \label{flowchart}}
\end{figure}

Figure~\ref{flowchart} presents a view of the algorithm.  The text to
be tagged is pre-processed and tokenized;
then morphological analysis\index{morphological analysis}
is performed followed by deterministic rules\index{deterministic rules}
and statistical knowledge\index{statistical knowledge}.
Finally, the large tagset\index{tagset} is reduced to a smaller one, as
an example of tag reduction.  Linguistic knowledge, statistical
learning, and tagset reduction modules are surrounded by dashed boxes
indicating that modules can be applied in arbitrary
order. Figure~\ref{flowchart} shows also an example of a few words in
a text for each of the modules in order to demonstrate the
disambiguation process.

The next sections describe the development of the tagger\index{tagger} based on
genotypes\index{genotype} estimation; first, some issues related to tokenization are raised, then
the linguistic knowledge and the statistical learning modules are
explained followed by the results.

\subsection{Issues with Tokenization}
The first step in tagging involves a series of text preprocessing modules
that are used for the tokenization of the corpus.

\begin{itemize}
\item {\bf Sentence boundaries:} places where sentences begin are
identified and replaced by appropriate tags. As punctuation
symbols play an important role in disambiguation, they are also
assigned special tags.
\item {\bf Proper nouns\index{proper nouns}:}
the morphological dictionary contains common nouns and proper nouns\index{proper
nouns}, but
there is a large number of proper nouns\index{proper nouns} that appear in the corpus and need to be
tagged; the number of proper nouns\index{proper nouns} missing in the morphological
dictionary is typically
fairly high.  Therefore, the tagger applies several heuristics. As an
example, it treats each word starting a sentence as possibly having an
additional {\it proper noun} tag; after morphological analysis, if the word
inherits a new analysis, the latter one will prevail; if not, the word is
identified as  {\it proper noun} and is 
dynamically added to the PROPER\_NOUNS dictionary.   If a capitalized
word is found in the middle of a sentence, it will inherit immediately
the {\it proper noun} tag.
\item {\bf Accent\index{French accents} restitution:}
An additional difficulty due to the accents\index{French accents} appears.  In continental French,
accented characters lose their accents\index{French accents} if they become capitalized. 
This is valid in either
sentence-initial position or in the middle of the sentence.  Therefore,
many accents in the text are missing.  A  phonology-based  recovery
technique is applied in order to
attempt to recover these accents\index{French accents}.
Namely, an initial uppercase vowel will get an accent if it precedes a
consonant in the following configuration:
\begin{itemize}
\item if the word starts with the following pattern {\tt ECV}, where
{\tt E} is the upper case character ``{\tt E}'', {\tt C} is one of the
consonants or consonant pairs
[b, bl, br, c, ch, cl, cr, d, dl, dr, f, fl, fr, g, gl, gr, h, j,
j, l, m, n, p, ph, pl, pr, q, r, s, sl, sr, t, tl, tr, v, vl, vr, z],
and V one the vowels [a, e, i, o, u, y], the acute accent is recovered.
\item if the observed word is ``A'' or ``Etre'', the accent will be
respectively grave and circumflex.
\end{itemize}
\item {\bf Acronyms:}\index{acronyms}  a treatment similar to the one of the proper
nouns\index{proper nouns} is applied here.
\item {\bf Compound words:}\index{compound words} compound words
or non-compositional words
in French are to be tagged as a separate entity.  They are recognized
from our dictionary sources and are 
considered as a single lexical unit. For example, locutions such as ``a
priori'' ({\it a priori}),
``top secret'' ({\it top secret}), or ``raz de mar\'{e}e'' ({\it tidal
wave}) will be treated as single lexical entries.
\item {\bf Personal pronouns: } if two words are connected by a dash
``-'', and the second word
is a personal pronoun, the two-word unit gets split.  For example, the
compound ``dit-elle'' ({\it said she}) becomes the two words ``dit'' and ``elle''.

\item {\bf Word splitting: } when all other stages are completed, the
corpus is split into lexemes and translated from 8-bit characters to
7-bit ascii characters if necessary.  Accents are marked with diacritic symbols
following the accented letter. Example: ``co\^{ }te's'' is used for
``c\^{o}t\'{e}s'' ({\it sides})\footnote{It is important to notice
that this marking does not introduce any ambiguity with the French
apostrophe, since apostrophes always occur after a consonant
whereas the accent marks always occur after vowels.}.
\end{itemize}

\section{Linguistic knowledge}
 Once the text is tokenized, morphological analysis is performed in order
to disambiguate the words.

\subsection{Morphological analysis}
Finite-state transducers\index{finite-state transducers}
(FST) are used to achieve morphological
analysis.  The FST\index{finite-state transducers}
is built on the model developed for Spanish
morphology \cite{TzouLib90} and handles mainly inflectional morphology
with some derivational affixes, such as ``anti-'' in ``anti-iranien''
({\it anti-iranien}), and ``arri\`{e}re-'' ({\it great}) in
``arri\`{e}re-grand-p\`{e}re'' ({\it great-grand-father}).  The
arclist dictionary -- dictionary of finite-state transitions -- was
originally built using several sources, including the Robert
Encyclopedic dictionary \cite{Robert} and lexical information from
unrestricted texts.

The FST\index{finite-state transducers}
used in the morphological stage of the tagger consisted of
up to 4 distinct sub-FST's: the common names or main
FST\index{finite-state transducers} , the main 
proper-noun FST, which is dynamically generated
from the learning corpus, and another proper-noun FST generated heuristically
from the corpus to be tagged or test corpus.  The last one is
generated each time a new corpus is tagged.
The main FST\index{finite-state transducers} recognizes
over 90,000 entries, i.e.  all inflected forms, such as nouns, verbs,
adjectives, as well as uninflected forms, such as adverbs, conjunctions,
and other categories.  Morphological analysis\index{morphological analysis}
is performed with a high
level of refinement.  For example, in addition to verbal forms
inflected for mood, tense, person, and number, pronouns are analyzed
into several categories, such as direct, indirect, disjoint,
reflexive, and so on.

\subsection{From features to tags} \label{tagset}
Once the morphological analysis\index{morphological analysis}
is performed, one needs to translate the
feature analysis into tags\index{tags}.  We use an abbreviation of the 
features of the word as its tag.  For example, the tag
{\tt BD3S} stands for a third person (3) singular (S) personal pronoun (B) 
direct object (D).

This offers several advantages:  first, it allows organization of the 
different categories by their syntactic feature, i.e. {\tt verb, noun}, etc; 
second, the tag reflects an interesting feature hierarchy.  For 
example, {\tt VIP3S} which is third person present indicative verb,
can be viewed in a feature hierarchy representation where verb is on top
of mood, tense, number, and person.
Third and consequently, rule operations can be done on any part of the
structure hierarchy.  For example, one can express generalizations 
on the tag paradigm, which simplifies the rule writing. In the
following example, one can replace the tense by a
metacharacter $[*]$: [{\tt
V3SPI,V3SFI,V3SSI,V3SII}] $\Rightarrow$ {\tt V3S*I}.  The rule will
apply to every verb in the indicative mood (I), for every tense (*)
which is in the third person (3) singular (S).
The first set of tags represents the detailed morphological analysis; 
it corresponds to
the large set of tags, i.e. 253 tags.   
Natural language systems, depending on what they try to achieve,
vary in the number of tags\index{tags} they require as well as in the choice of tags.
To address this issue, we left flexibility for the user so that
any set or subset of tags that is desired in connection with
the particular task in hand can be defined.  
The large set of tags\index{tagset} can be redefined by any subset of the same tag(s)
using a many-to-one mapping mechanism.
In our current tagging scheme, the 253 tags are collapsed at the end of the
tagging process to form a smaller set of 67 tags. 

\subsection{Negative constraints}
Linguistic knowledge\index{linguistic knowledge}
has been integrated in the system in the form of
negative rules\index{negative rules}.  
Several transformational rules specify for bigrams, trigrams, and larger
{\it n}-gram\index{n-gram} units that a particular sequence
of tags is not legal for a French sentence.
These rules are tightly dependent on
morphological analysis\index{morphological analysis}.
For example, the following negative constraints\index{negative constraints}
that list two continuous tags not admitted in French\index{French}, are introduced:

\begin{itemize}

\item {\bf BS3 BI1}. A BS3 (3rd person subject personal pronoun)
cannot be followed by a BI1 (1st person indirect personal pronoun).  
In the example: ``il nous faut'' ({\it we need}) -- 
``il'' has the tag BS3MS and ``nous'' has the tags [BD1P BI1P BJ1P BR1P BS1P].
The negative constraint ``BS3 BI1'' rules out ``BI1P'',
and thus leaves only 4 alternatives for the word ``nous''.

\item {\bf N K}. The tag N (noun) cannot be followed by a tag K
(interrogative pronoun); an example in the test corpus would be:
``... fleuve qui ...'' (...river, that...). Since ``qui'' can be
tagged both as an ``E'' (relative pronoun) and a ``K'' (interrogative
pronoun), the ``E'' will be chosen by the tagger since an
interrogative pronoun cannot follow a noun (``N'').

\item {\bf R V}. A word tagged with R (article) cannot be followed by a
word tagged with V (verb): for example ``l' appelle'' (calls
him/her). The word ``appelle'' can only be a verb, but ``l''' can be
either an article or a personal pronoun.  Thus, the rule will
eliminate the article tag, giving preference to the pronoun.
\end{itemize}

Negative constraints\index{negative constraints}
are examples of deterministic knowledge\index{deterministic knowledge}
introduced in the system. They express linguistic relationships
between the features of the words in a given {\it n}-gram, therefore
performing some contextual diambiguation over word strings.  These
relationships could perhaps be discovered through statistical
procedures, but since they are available to the human without
significant effort, they are easy to implement.
Each of the linguistic constraints\index{linguistic constraints}
is applied several times over the
words that have only one tag.  This iterative filtering process
generates words with unique tags, which serve as anchors in the
corpus.  In this incremental fashion, anchors can create new anchors
and thus enlarge the islands of disambiguated words.

\section{Statistical Learning\index{statistical learning}}
We manually tagged a set of three corpora, containing 10,000, 30,000,
and 36,000 words respectively, one from the ECI corpus -- extracted
from the newspaper ``Le Monde''--, and two others from other news articles.
Two additional corpora of 1,000 and 1,500 words were tagged
for testing purposes.  The test corpora were extracted from both
sources to reflect the two different text styles.

A statistical model based on {\it n}-gram probabilities was implemented
to find the best tag candidate for a given genotype\index{genotype}.
If {\bf t} is a tag and {\bf T} a tag genotype\index{genotype}, the question
is to find  $P(t|T)$, so that the most likely tag for a given word
can be selected.
Bigram\index{probabilities!bigram} probabilities were computed in estimating the
sequence of two tags given the two genotypes\index{genotype}, i.e.
$P(t_{i},t_{i+1}|T_{i},T_{i+1})$  and trigram\index{probabilities!trigram}
probabilities, i.e.
$P(t_{i},t_{i+1},t_{i+2}|T_{i},T_{i+1},T_{i+2})$.
Notice here that for bigrams and trigrams, the model does not estimate a single
tag occurrence but the sequence of tags.

\small
\begin{table}[htbp]
\centering
\begin{tabular}{|l|c|c|c|}	\hline
{\bf genotype}\index{genotype}		& {\bf  decision}	& {\bf freq. $f/n$}
& {\bf strength}\\ \hline
NMP P				& P		& 82/82		& 98.54
	\\ \hline
BD3S NMS RDF			& RDF 		& 172/173	& 98.44
	\\ \hline
BD3S RDM			& RDM		& 195/199	& 96.70
	\\ \hline
DMS NMS NXP RIMS W		& RIMS		& 107/109	& 96.30
	\\ \hline
P RP				& P		& 768/793	& 96.16
	\\ \hline
NMS pMS				& pMS		& 30/30		& 96.09
	\\ \hline
NXP W				& W 		& 90/92		& 95.63
	\\ \hline
NMP V2SPI V2SPS			& NMP 		& 25/25		& 95.33 
 \\ \hline
\end{tabular}
\caption{Best decisions that can be made according to unigram distributions. \label{top}}
\end{table}
\normalsize

Table~\ref{top} shows the best decisions that were made with {\it
n}-gram\index{probabilities!n-gram} probabilities.
For a given genotype\index{genotype} (1st column), the decision that was made 
over the 10,000 words training corpus (2nd column), the frequency of
this case occurence (3rd column),
and the strength of the decision (4th column) as explained below.

We use a $strength$ score for each statistical rule based
on the frequency, $f$, of the decision among $n$
observations of the tag genotype\index{genotype}. For instance, Table~\ref{top} gives $f=195$ and
$n=199$ for the decision RDM from the tag genotype\index{genotype} [BD3S,RDM]. The strength
score assumes that $f$ results from a binomial distribution\index{binomial distribution}
$B(p,n)$. This is the distribution which results when $n$
independent trials are made, each having probability $p$ of the
decision (and probability $1-p$ of any other member of the tag genotype\index{genotype}).
We do not know $p$, but must make an estimate from the data. When
$\hat{p}$ is estimated as the proportion $f/n$ of the decision in the tag genotype\index{genotype},
then the theory of the binomial distribution\index{binomial distribution}
\cite{Moore89} gives :
\[ sd(\hat{p}) = \sqrt{p(1-p)/n} \]
We estimate 
\[ \hat{p} = \frac{f + 0.5}{n + 1} \]
so that neither $\hat{p}$ nor $(1-\hat{p})$ will be zero. This procedure is
explained in Box (1973). We can estimate the uncertainty 
of $\hat{p}$ by:
\[ \sqrt{\hat{p}(1-\hat{p})/n} \]
and we use the value
\[ strength = (\hat{p} - \sqrt{\frac{\hat{p}*(1-\hat{p})}{n}}) * 100\]
to quantify the strength of the decision. This score represents the
estimate of the probability less
the estimate of the uncertainty. Notice in the above table that $25/25$ has
a lower strength than $30/30$ which in turn has a lower strength that $82/82$.
The strength measure is designed to give lower values for the same $f/n$
the smaller $n$ is.

The interesting aspect in this model, as demonstrated in Section~\ref{lex-prob}, 
is that probabilities are computed only on the genotypes\index{genotype}, and not on the words.
In using only unigram probabilities after the application of 
negative rules\index{negative rules}, the system disambiguates over 91\% of the text.
In applying bigram probabilities, system performance goes up to 93\%.

\section{Implementation and results}
Each component of the system -- tokenization,
morphological analysis\index{morphological analysis}, deterministic
rules, and statistical learning\index{statistical learning} --  is implemented as a
stand-alone program to preserve the modularity of the system.  In this
fashion, it is possible to use all the modules in any desired order,
except for the preprocessing and the morphology modules that are
applied in a fixed order (see Figure~\ref{flowchart} for a representation of the system).
The system can be viewed as a series of operators, each of which
performing some level of disambiguation of the morphological analysis.
In the following sections, operators or modules are studied in
different orders, so that one can scientifically test the relevance of
the best operator order.  The final output of the system contains all
words from the original corpus grouped in three categories: (1) the
correctly tagged words, (2) the incorrectly tagged words, (3) and
those that are still ambiguous.  The last group is particularly
important; it means that the evidence for disambiguating a certain
word is not sufficient at this point.  When tagging text relies on the
availability of training corpora but the amount of tagged text is
small, leaving words without a decision seems to be better than making
a decision without a strong enough level of confidence.  Moreover,
human taggers\index{taggers} do not always agree with one another, and it gives the
user the choice of picking the desired one.

  We used the system modularity and combined the operators in many
different ways.  This resulted in several experiments to figure out
the best path or the best order of module applications.  In varying
the parameters of the different modules as well as the ordering of the
modules, a total of 43 plausible tagging schemes was considered,
testing different orderings of (a) the deterministic stage, (b) the
statistical learning with different confidence thresholds\footnote{We rank
all possible unigram decisions according to their strength. In the different
tagging schemes, we vary the strength threshold in order to achieve optimal
results.}, (c) the application of unigram decisions, (d) and the tagset
reduction. 

\subsection{Previous experiments}
Three experiments were achieved in the previous version of this paper
\cite{TzouRadGal95a}.  The first one explored the modular aspect of the
system; several tagging combinations were performed in order to figure
out which modular path gives the best results.
Applying the operators in different order presents variants of the
correct/in\-cor\-rect/am\-bi\-guous tagging path.  The scheme that achieves
the largest percentage of correct tags\index{tags} is the one that applies sequentially
Morphology\index{morphology} (M), Negative Constraints\index{negative constraints}
(with 3 iterations) (D),
Statistical Decisions\index{statistical learning}
with maximal coverage ($A_{5}$), and Tag Reduction\index{tagset} (T).
At this point, the system performance was 90.4\% correct, 8.4\% incorrect, and 
1.2\% ambiguous.
At each iteration during the deterministic stage, the system tries to find anchors
(i.e. words, which at that moment have been disambiguated). These words are
used to propagate negative constraints\index{negative constraints}
to their neighbors.
We found empirically that after 3 iterations, the number of anchors does not increase
due to the large number of inherently unambiguous words in the texts.

In the second experiment, we varied the threshold so that we
could analyze the effect on the performance.  It turned out that a
lower value of the threshold represents more 
(90.4\% correct) but possibly incorrect (8.4\% incorrect) statistical
decisions, whereas a
higher value gave fewer (83.4\% correct) but more reliable decisions
(3.9\% incorrect).  The last one explored the impact of the two
different tagsets on the tagger performance.  Interestingly, the
reduction from the large tagset to the small one does not improve much
(only .09\%) the performance of the tagger.  This is mainly due to
the fact that the reduction is done mainly inside the main
part-of-speech categories (i.e. verbs, nouns, etc.)  and not accross
the categories.

More recently, we have been working on re-implementing the part-of-speech
tagger\index{tagger} using only a finite-state machine\index{finite-state transducer}
framework. We have used a toolkit developed by Pereira et al. (1994)
which manipulates weighted and unweighted finite-state machines (acceptors
or transducers). 
Using these tools,  we have created a set of programs
which generate finite-state transducers\index{finite-state transducers} from descriptions of
negative constraints\index{negative constraints}, as well as other transducers for
the statistical learning\index{statistical learning}.
Statistical decisions on genotypes\index{genotype} are
represented by weights. Negative constraints\index{negative constraints}
are assigned the highest
cost. In the earlier framework, conflicting {\it n}-gram decisions were handled
in an arbitrary fashion - choosing the ones closer to an anchor. With the
finite-state tools, we are able to prefer one {\it n}-gram decision over another
based on their cost.
For example, in order to trace the disambiguation stages, the word
``moyenne'' used in Table~\ref{marine} was observed (genotype\index{genotype} and
resolution are represented in bold characters).
Table~\ref{gen-res} contains three steps of resolution.  The first
one, unigram probabilities\index{probabilities!unigram}, exhibits the genotype\index{genotype} in the left column
with the resolution in the second column and the frequency in third
one.  At this point, the verb tag ``v3s'' is in first position (with
41.54\%).  The second one, bigram probabilities\index{probabilities!bigram},
shows the first two
genotypes\index{genotype} providing a score of 100\% and 75\% in favor of the tag ``jfs'' for
the word ``moyenne''. The third line indicates that for the bigram
genotype\index{genotype} [[nms],[jfs nfs v1s v2s v3s]], the resolution is 100\% in favor of ``v3s''.
In the context of a preceeding word being a masculine singular noun (``nms''), 
it is understandable that the verb tag (``v3s'') is more likely to be correct.
The trigram probabilities\index{probabilities!trigram} have two resolutions,
one with a masculine noun on the left (``nms''), the other with a feminine noun
also on the left (``nfs''), and the two cases still favor the adjective tag
``jfs'' for ``moyenne''.   It shows that if the preceeding word is a noun,
no matter its gender, the genotype\index{genotype} [jfs nfs v1s v2s v3s] will be resolved by
picking the adjective tag (``jfs'').  This is another persuasive example of the
power of {\it n}-gram genotype\index{genotype} resolution that doesn't require lexical
probabilities. 

\begin{table}[htbp]
\centering
\begin{tabular}{|l|l|l|}	\hline
\multicolumn{3}{|c|}{\bf unigram probabilities}\\\hline
{\bf genotype}\index{genotype} & {\bf resolution} & {\bf prob.} \\\hline
{\bf [jfs nfs v1s v2s v3s]}   & {\bf v3s}  &   41.54\%   \\
{\bf [jfs nfs v1s v2s v3s]}   & {\bf jfs}  &   35.38\%   \\
{\bf [jfs nfs v1s v2s v3s]}   & {\bf nfs}  &   23.08\%   \\\hline \hline
\multicolumn{3}{|c|}{\bf left bigram probabilities} \\\hline
{\bf bigram genotype} & {\bf resolution}          &   {\bf prob.} \\\hline
[nfs], {\bf [jfs nfs v1s v2s v3s]} & [nfs, {\bf jfs}] & 100.00\% \\
$[$nfs nms$]$, {\bf [jfs nfs v1s v2s v3s]} & [nfs, {\bf jfs}] & 75.00\% \\
$[$nms$]$, {\bf [jfs nfs v1s v2s v3s]} & [nms, {\bf v3s}] & 100.00\% \\ \hline\hline
\multicolumn{3}{|c|}{\bf trigram probabilities} \\\hline
{\bf trigram genotype} & {\bf resolution} &  {\bf prob.} \\\hline
[nfs nms], {\bf [jfs nfs v1s v2s v3s]}, [a b p] & [nms, {\bf jfs}, p] & 50.00\%   \\
$[$nfs nms$]$, {\bf [jfs nfs v1s v2s v3s]}, [a b p] & [nfs, {\bf jfs}, p] &
50.00\%   \\\hline
\end{tabular}
\caption{An example of genotype\index{genotype} resolution. \label{gen-res}}
\end{table}

Table~\ref{res} presents the current performance of the tagger
according to the different
{\it n}-gram probabilities\index{probabilities!n-gram} and the application of the linguistic rules.  
The results are based on the small training corpus of 10,000 words and we believe
the performance will get better in using the entire training corpus. 
The figures shown in Table~\ref{res} reflect the percentage of words that are
disambiguated correctly by the tagger. The remainder to 100\% consists of
words that have been incorrectly disambiguated.

\begin{table}[htbp]
\centering
\begin{tabular}{|l|l|l|}	\hline
    &  {\bf  unigrams}  &  {\bf bigrams}\\\hline
{\bf 10K-word training corpus} & 92.0\% & 93.0\%  \\\hline
\end{tabular}
\caption{Tagger performance with {\it n}-gram probabilities and
negative constraints. \label{res}}
\end{table}

\subsubsection{Comparative study of the two tagsets \label{comp}}

The goal in building a flexible tagset\index{tagset} is to allow the user to pick
whatever set is necessary for a particular application.  Even though
the reduction in size from the large to the small set is of 74\%, this
gain was not reflected in the tagged text.  In order to understand
better this phenomenon, we observed the distribution of tags in a
corpus of over 200,000 words.  We collected the frequencies of the
tags\index{tags} in the two different sets and the top
35 most frequent tags\index{tags} appear in Table~\ref{freq}.
The large tagset represents the
morphological features obtained by the morphological analyzer.  In
constructing the small set, we eliminated a large number of
morphological features that are of relatively little use at the
syntactic level; for example, mood and tense for verbs, reflective,
disjoint, and subject position for personal pronouns.
Additionally, auxiliaries and verbs -- a total of 93 different forms
-- were collapsed in fewer categories and only the person and the
number were kept, which resulted in a total of 13 different tags.  All
personal pronouns were collapsed and the numbers went from 79 to 9.
Column 2 of Table~\ref{freq} shows the large tagset\index{tagset} with the number of
tag occurrences (column 1), and the tag meaning (column 3).  Column 4,
5, 6 show similar information for the small tagset\index{tagset}.
As an example, we
highlighted the different occurences of third person singular verbs on
the left of the table that correspond to a single tag occurence in the
small tagset on the right side.

\tiny
\begin{table}[htbp]
\centering
\begin{tabular}{|l|l|l||l|l|l|}	\hline
{\bf Num.}  &{\bf Large} &                   & {\bf Num.}  &   {\bf Short}       &  \\ 
{\bf of occ.} &{\bf Tagset}& {\bf Description} & {\bf of occ.} & {\bf Tagset} & {\bf Description} \\ \hline
31562 & P 	  & prep.                   & 31562 & p           & prep.             \\
19567 & . 	  & punct.                  & 19567 & x           & punct. 	       \\
12398 & NFS 	  & fem. sg. n.           & 12398 & nfs         & fem. sg. n. 	       \\
11792 & NMS 	  & masc. sg. n.          & 11792 & nms         & masc. sg. n. 	       \\
8650 & A 	  & adv.                  &  8650 & a           & adv. 	               \\
7445 & U 	  & proper n.             &  8169 &{\bf v3s}    & {\bf 3rd sg.}        \\
     &            &                       &       &             & {\bf  verb/aux.} \\
7375 & RDF 	  & fem. def. art.          &  7445 & u           & proper n. 	       \\
6975 & $\hat{}$   & beg. of sent.           &  7375 & rf          & fem. def. art.       \\
6975 & \$ 	  & end of sent.            &  7074 & r           & indef. art.          \\
4631 & W 	  & numeral                 &  6975 &$\hat{}$     & beg. of sent.   \\
4467 & {\bf V3SPI}& {\bf 3rd sg.}           &  6975 & \$ & end of sent. \\
     &            & {\bf pres. ind. verb}   &       &    &                \\
4363 & NMP 	  & masc. pl. n.                       &  6208 & b & pers. pron.         \\
4171 & CC 	  & coord. conj.                         &  4631 & z & numeral            \\
4002 & i 	  & inf. verb                        &  4444 & v & inf. verb/aux..   \\
3958 & NFP 	  & fem. pl. n.                        &  4363 & nmp & masc. pl. n.      \\
3726 & RDM 	  & masc. def. art.                     &  4171 & cc & coord. conj.  	   \\
3471 & QSMS 	  & masc. sg. past part.            &  3958 & nfp & fem. pl. n.            \\
3379 & JMS 	  & masc. sg. adj.                       &  3910 & qsms & masc. sg.       \\
     &            &                                      &       &      & past part.\\
3299 & RDP 	  & partitive art.                      &  3726 & rm & masc. def. art. 	     \\
2883 & JXS 	  & sg. inv. (gender) adj.             &  3379 & jms  & masc. sg. adj.        \\
2597 & CS 	  & subord. conj.                        &  3275 & js & sg. adj. 	   \\
2563 & NMX 	  & masc. inv. (number) n.           &  2898 &v3p   & 3rd pl. verb       \\
2469 & JFS 	  & fem. sg. adj.                        &  2597 & cs   & subord. conj.      \\
1995 & RIMS 	  & masc. sg. indef. art.             &  2571 & nm   & masc. n.               \\
1979 & JMP 	  & masc. pl. adj.                       &  2469 & jfs  & fem. sg. adj.       \\
1743 & E 	  & rel. pron.                         &  1979 &jmp   & masc. pl. adj.         \\
1739 & RIFS 	  & fem. sg. indef. art.              &  1976 & jp   & pl. adj.               \\
 1511 & BS3MS 	  & 3rd sg. subj. pers. pron.      &  1618 & bms  & masc. pers. \\
     &            &                                &       &      & pron.\\
1449 & {\bf \&3SPI}& {\bf 3rd sg. pres.}           &  1251 & jfp & fem. pl. adj.        \\
     &             & {\bf ind. aux.}               &       &     &                      \\
1392 & BR3S 	  & 3rd sg. refl. pers. pron.      &  1088 & jms  & masc. sg. adj.           \\
1319 & JXP 	  & pl. inv. (gender) adj.             &   876 & qp   & pres. part.         \\
1251 & JFP 	  & fem. pl. adj.                        &   811 & qsmp & masc. sg.          \\
     &            &                                      &       &      & past part.\\
1129 & V3PPI 	  & 3rd pl. pres. ind. verb      &   746 & h    & acronym              \\
1063 & \&3PPI 	  & 3rd pl. pres. ind. aux.    &   744 & qsfs   & fem. sg.           \\ 
     &            &                                      &       &      & past part.\\\hline
\end{tabular}
\normalsize	
\caption{Most frequent 35 tags in a corpus of 200,000 words for large and small tagsets. \label{freq}}
\end{table}
\normalsize

Any subset of this large tagset\index{tagset} can be
(re)defined by the user with a very simple mapping mechanism.  This is an
important feature of the system design since it makes the tagger adaptable
to different NLP applications\index{NLP applications}
requiring different sets of tags\index{tags} or morphological variants.

\section{Related Research}
A number of taggers\index{taggers} and tagging methods are available.
For the last decades,
 part of speech tagging systems have generally followed either a rule-based approach
\cite{KleinSimm63}, \cite{Brill92}, \cite{Voutilai93}, or a statistical one
\cite{BahlMercer76}, \cite{LeeGarAtw83}, \cite{Merialdo94},
\cite{DeRose88}, \cite{Church89b}, \cite{CuttKupiec92}.
 Statistical approaches often use  Hidden Markov Models for estimating
lexical\index{lexical probabilities} and
contextual probabilities\index{contextual probabilities},
while rule-based systems capture linguistic generalities
to express contextual rules.
Most of these approaches have benefited from large tagged corpora mentioned
above,  which make the training and testing procedures feasible.

Brill and Marcus (1992) and Brill (1992) proposed a simple and powerful
corpus-based language modeling approach that learns a series of
transformational rules that are then applied in sequence to a test corpus
to produce predictions.  The learning approach combines a large training
corpus, a baseline heuristic for selecting initial default values, and a
set of rule templates defining classes of transformational rules that use
particular neighborhood characteristics as the grounds for changing a
particular current value.  For example, in the part of speech tagging
application, the baseline heuristic might be to assign to each ambiguous
word whatever tag is most often correct in the training corpus, and the
templates, defined here over a window that includes a context of two tags
on each side, will apply the rules if the tag transition needs to be
changed.   It seems clear, though, that the
performance of Brill's tagger is contingent on the availability of a large
training corpus. Since rules are acquired automatically, a small training
corpus cannot provide enough empirical data for the acquisition of a large
number of rules.

Chanod and Tapanainen (1995) compare two tagging frameworks for French, one
that is statistical, built on the Xerox tagger \cite{CuttKupiec92},
and another based on linguistic constraints only.  The constraints can
be 100\% accurate or describe the tendency of a particular tagging
choice.  The constraint-based tagger is proven to have better
performance than the statistical one, since rule writing is more
handlable or more controllable than adjusting the parameters of the
statistical tagger.  The tagset used is very small (37 tags),
including a number of word-specific tags (which reduces further the
number of tags), and does not account for several morphological
features, such as gender, number for pronouns, etc.  Moreover,
categories that can be very ambiguous, such as coordinating
conjunctions, subordinating conjunctions, relative and interrogative
pronouns tend to be collapsed; consequently, the disambiguation
problem is too simplified (therefore giving high performance) and
results cannot be compared since the ambiguities do not lie at the
same word level.

Merialdo (1994) makes comparisons among different tagging
schemes using classic Viterbi algorithms on the one hand, and Maximum
Likelihood Estimation tagging on the other.  In \cite{Merialdo94}, results
show that the estimation of the model parameters counting the relative
frequencies of a large quantity of hand-tagged corpus gives better
results than training using Maximum Likelihood.

To contrast with these different approaches, our work has attempted to
go in more nuances and refinements of the linguistic subtleties.
Therefore, the disambiguating task is rendered more complex.  In the
spectrum of parts of speech taggers as described above, the
originality of our work lies in the linguistic nuances and subtleties
that are encoded in the system.
Part of speech taggers\index{part-of-speech taggers} could be
divided in two categories, these which discriminate simple part of
speech categories, such as \cite{Church89b} and \cite{ChanodTap95} and
those like \cite{Voutilai93}, which detect noun phrases.  Like
Voutilainen's system, ours provides more morpho-syntactic information
and therefore tackles more of the linguistic ambiguities.
With its numerous encoded morphological features, not only is it more
flexible to different NLP applications, but it can be also viewed as
the first step towards a shallow parsing system.

\section{Remarks and Future Work}
This paper presents techniques for assigning the most appropriate tag\index{tag}
among all the ones generated by the
morphological analysis\index{morphological analysis} for each
French\index{French} word in unrestricted texts.  We explored different strategies
to capture both morphological and syntactic variants.  With a
restricted amount of tagged data,
lexical probabilities\index{lexical probabilities} were shown to
be limited in their predictive ability.  Even large amounts of
training data would not solve the problem of sparseness\index{sparse data}
in training data.  The solution to this problem was to select the set of tags
associated with a word -- the genotype\index{genotype} -- and
apply linguistic knowledge\index{linguistic knowledge}
and statistical learning\index{statistical learning} on this unit.  This approach
exhibits also an elegant way of smoothing probabilities, in giving
estimates of unseen words in tagging.  A tagger for unrestricted text
was then built that took into account the limitation of training data,
and combined empirical and symbolic methods to disambiguate word parts
of speech.  Among others, the contribution of this work resides in the
successful approach of using the genotype\index{genotype} to
estimate {\it n}-gram probabilities\index{probabilities!n-gram}.
We are in the process of improving the system
performance and are exploring the portability of the system to other
languages, such as Spanish.

\noindent
{\bf Acknowledgments}\\ We wish to thank Ido Dagan, Judith Klavans, and
Diane Lambert for their valuable comments throughout this work.

\printindex
\end{document}